\documentclass[preprint,12pt]{elsarticle}

\usepackage{amssymb}
\usepackage{amsmath}

\usepackage{lineno}
\usepackage{natbib}
\usepackage{algorithm}
\usepackage{algorithmic}
\usepackage{graphicx}
\usepackage{subfig}
\usepackage{url}
\usepackage{booktabs}
\newtheorem{statement}{Statement}
\usepackage{newfloat}
\usepackage{multirow}

\journal{Nuclear Physics B}

\begin{document}

\begin{frontmatter}



\title{CausalCOMRL: Context-Based Offline Meta-Reinforcement Learning with Causal Representation}
\author[label1]{Zhengzhe Zhang}
\ead{zhengzhezhang@mail.sdu.edu.cn}
\author[label1]{Wenjia Meng\corref{cor1}}
\ead{wjmeng@sdu.edu.cn}
\author[label1]{Haoliang Sun}
\ead{haolsun.cn@gmail.com}
\author[label3]{Gang Pan}
\ead{gpan@zju.edu.cn}

\cortext[cor1]{Corresponding author.}
\affiliation[label1]{organization={School of Software},
            addressline={Shandong University},
            city={Jinan},
            postcode={250101},
            country={China}}
\affiliation[label3]{organization={The State Key Lab of Brain-Machine Intelligence},
            addressline={Zhejiang University},
            city={Hangzhou},
            postcode={310027},
            country={China}}

\begin{abstract}
Context-based offline meta-reinforcement learning (OMRL) methods have achieved appealing success by leveraging pre-collected offline datasets to develop task representations that guide policy learning. However, current context-based OMRL methods often introduce spurious correlations, where task components are incorrectly correlated due to confounders. These correlations can degrade policy performance when the confounders in the test task differ from those in the training task. To address this problem, we propose CausalCOMRL, a context-based OMRL method that integrates causal representation learning. This approach uncovers causal relationships among the task components and incorporates the causal relationships into task representations, enhancing the generalizability of RL agents. We further improve the distinction of task representations from different tasks by using mutual information optimization and contrastive learning. Utilizing these causal task representations, we employ SAC to optimize policies on meta-RL benchmarks. Experimental results show that CausalCOMRL achieves better performance than other methods on most benchmarks.
\end{abstract}




\end{frontmatter}


\section{Introduction}
Deep reinforcement learning (DRL) has demonstrated significant success across various domains, including robotics~\cite{kober2013reinforcement,peng2018sim,wang2021modular}, gaming~\cite{mnih2015human,silver2018general,patel2019improved}, and embodied artificial intelligence~\cite{tobin2017domain,robertazzi2022brain,kaadoud2022explaining}.
However, the practical application of DRL faces significant challenges in data efficiency and generalization. 
These challenges require extensive online interactions between the agent and environment to develop effective policies~\cite{chua2018deep}, which restricts the application of deep reinforcement learning to real-world tasks~\cite{zhao2022offline}. 
Offline reinforcement learning addresses the challenges of data efficiency and high costs of online interactions by leveraging pre-collected offline datasets~\cite{levine2020offline}. 
However, offline reinforcement learning has difficulties in generalizing to new scenarios when task dynamics and reward functions vary. 
This limitation often requires the collection of new task-specific datasets for effective adaptation and training.

To overcome the above limitation, the offline meta-reinforcement learning (OMRL) method is proposed, which trains on a variety of similar tasks and utilizes prior knowledge to rapidly adapt to new tasks. Offline meta-reinforcement learning methods can roughly be categorized into two main approaches: gradient-based OMRL methods~\cite{schweighofer2003meta,finn2017model,gupta2018meta,rengarajan2022enhanced} and context-based OMRL methods~\cite{rakelly2019efficient,li2020focal,ben2022context}. Gradient-based OMRL methods begin by initializing the policy model with predefined parameters and then enhance the model through a few gradient updates to enable quick adaptation. 
Context-based OMRL methods incorporate a task inference process that leverages historical samples to form a task representation~\cite{zhao2023representation}, guiding the policy learning process. 
Compared to gradient-based OMRL methods, context-based OMRL methods demonstrate greater resistance to negative transfer in multi-task learning~\cite{zhang2022survey} and achieve more efficient sample utilization~\cite{beck2023survey,zhou2023episodic}. 
\begin{figure*}[htb!]
  \centering
    \subfloat[Causal Graph Example]{
        \label{fig:a}
  		\includegraphics[width=0.5\textwidth]{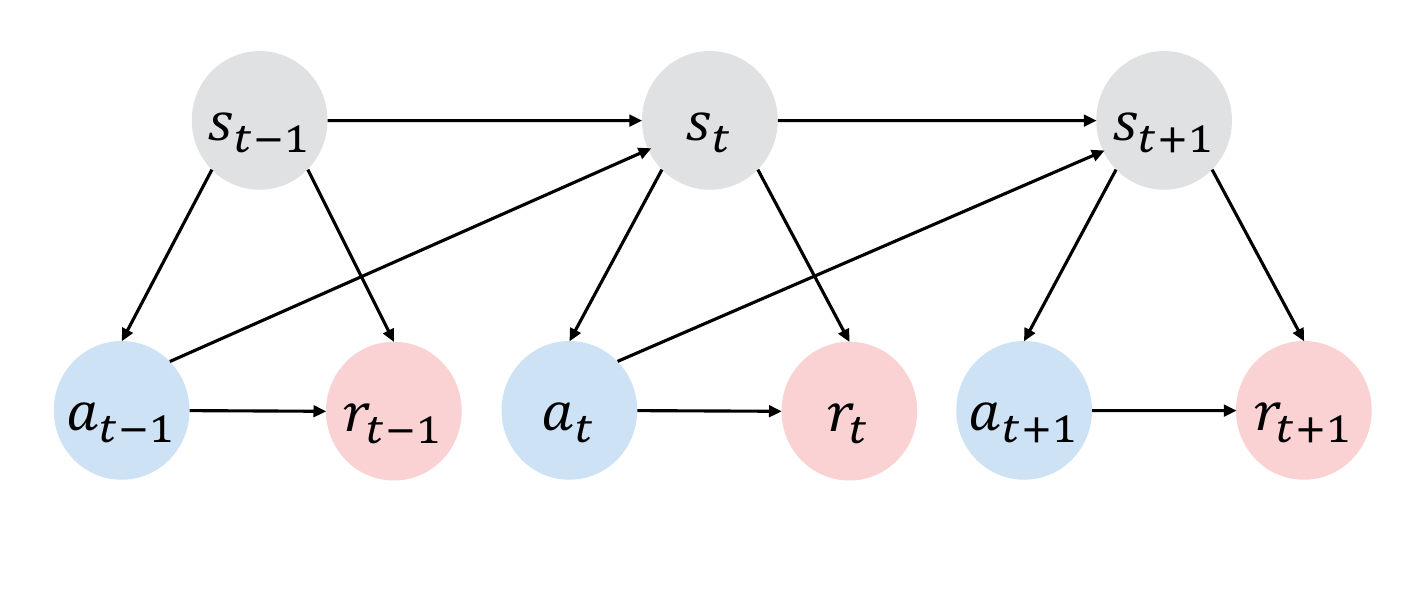}
  }
  \subfloat[Task Representation Embedding]{
        \label{fig:b}
  		\includegraphics[width=0.5\textwidth]{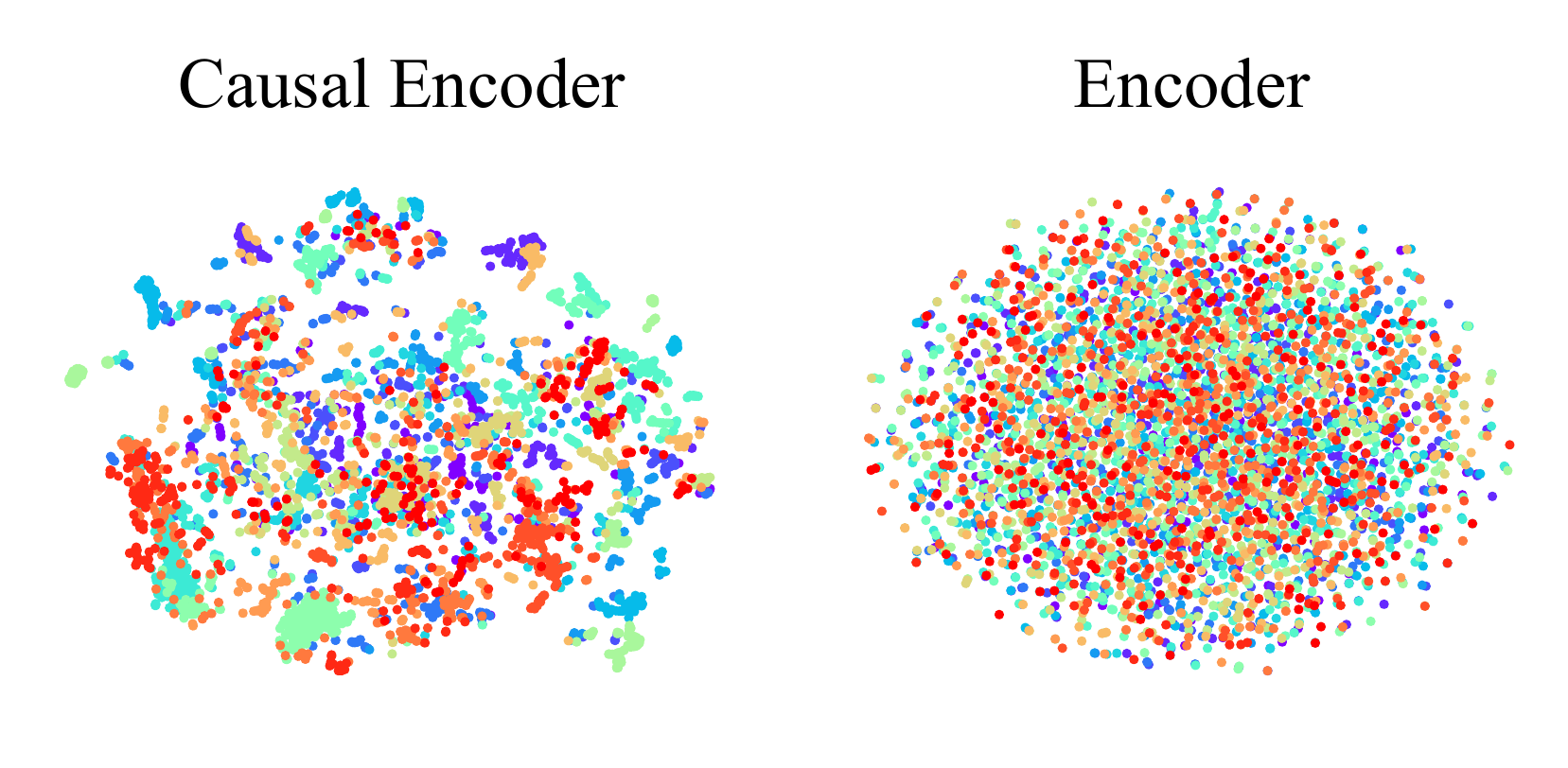}
  }
  \caption{Causal graph and encoder performance comparison. (a) Causal Graph Example. In the causal graph example, the nodes represent the state, action, and reward at varying timesteps, with edges indicating the causal relationships among them. $t$ represents the timestep. (b) t-SNE visualization of the task representation embedding vectors in Walker-Rand-Params. }
  \label{causal}
\end{figure*}

In context-based OMRL, it is crucial to learn task representations effectively. 
Existing work on task representation can be roughly classified into three categories: metric learning-based methods, contrastive learning-based methods, and mutual information-based methods. 
Metric learning-based methods~\cite{li2020focal,li2021provably} utilize metric learning to capture the structure of the task representation. Contrastive learning-based methods~\cite{yuan2022robust, wang2023meta} use contrastive learning to optimize task context encoder for learning the task representation. 
Mutual information-based methods~\cite{gao2024context, li2024towards} utilize mutual information optimization to learn the task representation. 
However, representation learning strategies in current methods often introduce spurious correlation, where different task components are mistakenly correlated due to the influence of confounders~\cite{ding2024seeing, hu2022improving}. 
Spurious correlations are common in reinforcement learning tasks. For example, in the \emph{Walker-Rand-Params}~\cite{todorov2012mujoco}, the robot received a high reward when the agent's top maintained a specific tilt angle, influenced by specific conditions such as an initial speed or balance state. 
A task representation that learns such useless or even harmful correlations could negatively impact policy performance when the confounders in the test task diverges from that in the training task~\cite{scholkopf2021toward}.

Causal representation learning leverages structured causal models, such as causal graphs, to identify the true relationships among variables~\cite{scholkopf2021toward}. 
This approach helps learn the causal effect of the state, action, and reward in an RL task, preventing misguidance by spurious correlations. 
Thus, it enables the agent to adapt more effectively to unseen tasks in meta-RL and enhance generalization~\cite{wang2021provably,scholkopf2021toward}. 
To clearly illustrate the causal structure in RL tasks and highlight the benefit of causal representation learning, we present a concrete example in Figure \ref{causal}. 
Figure \ref{causal}(a) depicts a possible causal graph for a meta-RL task, where nodes represent portions of the task and the edges denote the causal relationships between them. 
Figure \ref{causal}(b) presents the t-SNE visualization of the task representation, uniformly sampled for test tasks and displayed using red-to-purple points. Each task is uniquely identified by a specific color. 
Causal task representations exhibit tighter clustering for the same task and greater distinction between different tasks, indicating that causal representation learning helps the encoder better distinguish task contexts on test tasks.

In this paper, we propose a causal context-based OMRL method termed CausalCOMRL, which leverages causal representation learning. 
This approach learns the causal relationships among task components and integrates causality into the task representations, thereby enhancing the generalizability of context-based OMRL. 
Specifically, we utilize causal representation learning to learn the task representation that contains the causal relationship among task components. 
To enhance the task representation learning process, we further incorporate mutual information optimization and contrastive learning to increase the differentiation between representations of different tasks. 
With the above task representation, we use a representative reinforcement learning method (SAC~\cite{haarnoja2018soft}) to optimize policies on meta-RL benchmarks. 
Extensive experimental results conducted on representative benchmarks demonstrate the effectiveness of the proposed CausalCOMRL method.
Our contributions can be summarized as follows:
\begin{itemize}
  \item We propose a causal context-based offline meta-reinforcement learning method (CausalCOMRL) which integrates causal representation learning to optimize the task encoder and acquire causal relationships among task components. This is the first approach to incorporate causality into task representations within the context-based OMRL setting.
      
  \item To enhance the discrimination ability of the causal task encoder for various tasks, we introduce mutual information and contrastive learning into the optimization process of the task encoder in CausalCOMRL.
  
  \item The extensive experiments conducted on representative meta-RL benchmarks demonstrate that CausalCOMRL outperforms other context-based OMRL methods on most benchmarks. The visualizations of task representation also validate the effectiveness of our causal task encoder.
\end{itemize}

\section{Related Work}
\noindent\textbf{Offline Reinforcement Learning.} Unlike online reinforcement learning, which involves agent-environment interaction during training, offline reinforcement learning operates without such interaction, learning from fixed datasets collected by unknown behavior policies. Recent offline reinforcement learning methods~\cite{levine2020offline,nair2020awac,zhang2024perspective} focus on addressing distributional shifts and value overestimation due to discrepancies between target and behavior policies. Our work advances the field of offline reinforcement learning by emphasizing the learning of causal task representations and meta-policies from offline datasets.

\noindent\textbf{Offline Meta-Reinforcement Learning.} Offline meta-reinforcement learning (OMRL) aims to address the challenges of generalization and costly data collection by applying meta-learning techniques to pre-collected offline datasets. OMRL methods focus on learning task representation during the meta-training process by utilizing historical trajectories from offline datasets. 
Current OMRL methods are categorized into gradient-based and context-based approaches. Gradient-based methods \cite{finn2017model,gupta2018meta,rengarajan2022enhanced} optimize initial policy parameters for quick task adaptation. Context-based methods \cite{rakelly2019efficient,li2020focal,wang2023offline} frame tasks as contextual Markov Decision Processes (MDPs), focusing on encoding task representations from offline data.
Our work adheres to the context-based OMRL approach, emphasizing the learning of causal task representations to enhance the generalization.

\noindent\textbf{Causal Representation Learning.} 
The central problem for causal representation learning is to discover high-level causal variables from low-level observations~\cite{scholkopf2021toward}. Current causal representation learning methods mostly fall into two categories: the first category methods~\cite{ahuja2022weakly,brehmer2022weakly,gresele2021independent,lachapelle2022disentanglement} realize causal representation learning under supervision of ground truth counterfactual images generated according to causal graph; the second category methods~\cite{kocaoglu2017causalgan,yang2020causalvae,shen2022weakly,reddy2022causally} realize representation learning under annotations and causal graph. Although these methods have been widely applied in computer vision, the application of causal representation learning in reinforcement learning settings is still an open problem~\cite{seitzer2021causal,ding2022generalizing,lampinen2024passive}. In this paper, we apply causal representation learning in a context-based OMRL setting to enhance the generalization ability. To the best of our knowledge, this is the first work to propose the application of causal representation learning in the context-based OMRL setting.

\noindent\textbf{Mutual Information Optimization and Contrastive Learning.}
Mutual information optimization and contrastive learning are two critical techniques for learning data representation. Specifically, mutual information optimization learns the essential data representation by enhancing the correlation between variables and maximizing their mutual information~\cite{oord2018representation,mu2022domino, choshen2023contrabar}. Contrastive learning learns data representation by maximizing the similarity between related samples and minimizing the similarity between unrelated samples~\cite{chen2020improved,caron2021emerging,yan2021consert,gao2021simcse,liu2023contrastive, oord2018representation}. Our work introduces mutual information optimization and contrastive learning into task representation learning for context-based OMRL, aiming to enhance the representation learned through causal representation learning.

\section{Preliminaries}
\subsection{Problem Formulation}
Reinforcement learning (RL) \cite{sutton2018reinforcement} is modeled as a Markov Decision Process (MDP), $M=\left(S, A, P, R, \rho, \gamma\right)$, where $S$ denotes the state space, $A$ denotes the action space, $P(s'|s,a)$ is the transition probability of state $s'$ under taking action $a$ under state $s$, $R(s,a)$ denotes the reward function, $\rho(s)$ denotes the initial state distribution, and $\gamma \in (0,1)$ is the discount factor. 
A policy $\pi$ is a probability distribution defined on $S \times A$, and $\pi (a_t|s_t)$ denotes the probability of taking $a_t$ under state $s_t$ at timestep $t$. 
We define the state value function and the action value function as follows:
\begin{equation}
V_\pi\left(s_t\right)=\sum_{n=0}^{\infty}\gamma^n \mathbb{E}_{a_t \sim \pi (\cdot|s_t), s_{t+1} \sim P\left( \cdot \mid s_t, a_t\right)}\left[R\left(s_{t+n}, a_{t+n}\right)\right],
\end{equation}
\begin{equation}
Q_\pi\left(s_t, a_t\right)=R\left(s_t, a_t\right)+\gamma \mathbb{E}_{s_{t+1} \sim P\left(\cdot \mid s_t, a_t\right)}\left[V_\pi\left(s_{t+1}\right)\right].
\end{equation}

In this paper, we study context-based offline Meta-Reinforcement Learning (OMRL) \cite{rakelly2019efficient}, where the agent learns the policy from the offline datasets that are generated by the task distribution $p(M_i)$, where $M_i=\left(S, A, P_i, R_i, \rho, \gamma\right)$.
Different from standard RL, the rewards and transition dynamics in OMRL are task-varying, denoted as $ p(P,R)$. 
Let transitions $\{(s_{i,j}, a_{i,j}, s'_{i,j}, r_{i,j})\}^{N}_{j=1} =: \mathcal{D}_i $ be the offline dataset with size $N$, where $\mathcal{D}_i$ is generated according to the process $M_i$. Furthermore, the context-based OMRL encodes the context $c=:\{(s_{i,j'}, a_{i,j'}, s'_{i,j'}, r_{i,j'})\}^{n}_{j'=1}\subset \mathcal{D}_i$ to produce a task-specific latent representation $z$. 
The latent representation incorporates distinct task characteristics and facilitates the learning process of meta-policy $\pi_\theta (a|s,z)$ and the value functions conditioned on $z$.
The goal of context-based OMRL is to develop a meta-policy $\pi_{\psi}$ to maximize the expected cumulative reward:
\begin{equation}
	J\left(\pi_\theta\right) = \mathbb{E}_{M_i \sim p(\cdot)}\left[\mathbb{E}_{\tau_{i}}\left[\sum_{t=0}^{\infty} \gamma^t R\left(s_t, a_t\right)\right]\right],
\end{equation}
where $\tau_{i}$ is the transition trajectory sampled from task $M_i$.

\begin{figure*}[htb!]
	\centering
	\includegraphics[width=\textwidth]{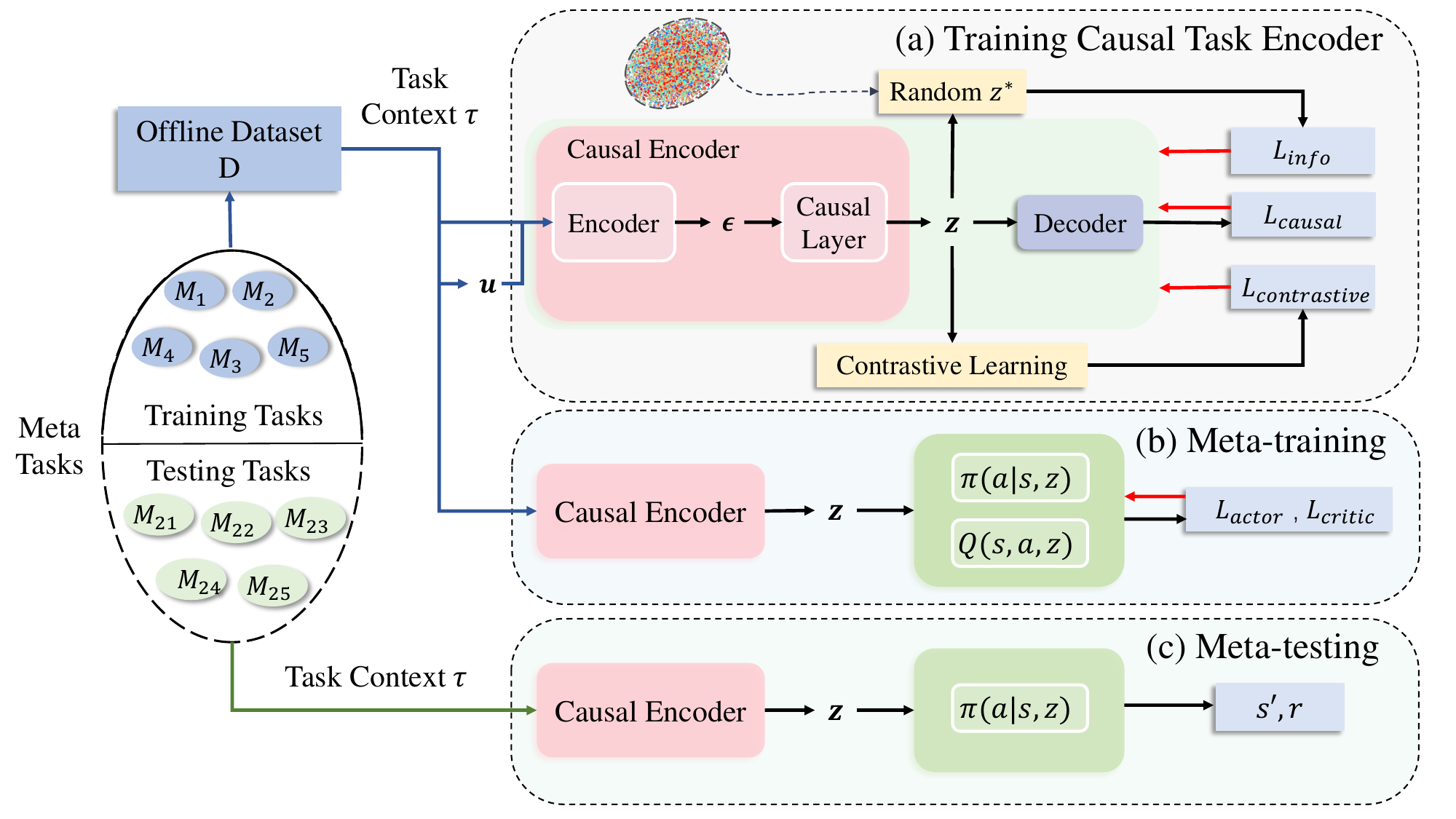}
	\caption{Framework of CausalCOMRL: (a) Causal task encoder training module. (b) Meta-training module. (c) Meta-testing module.}
	\label{framework}
\end{figure*}

\section{Proposed Method}
In this section, we propose CausalCOMRL, a context-based OMRL method that leverages cause-and-effect relationships in task representation to enhance structured knowledge learning and improve task generalization. 
We first describe the framework of CausalCOMRL, which consists of three key modules: causal task encoder training module, meta-training module, and meta-testing module. 
Next, we introduce a task encoder with causal representation learning and enhance it using contrastive learning and mutual information.
Finally, we detail the practical implementations of the proposed CausalCOMRL. 

\subsection{Framework of CausalCOMRL}
The CausalCOMRL framework consists of three modules: a causal task encoder training module, a meta-training module, and a meta-testing module, as shown in Figure \ref{framework}. In the following, we will introduce these modules in detail.

In causal task encoder training module: we start by extracting context trajectories $\tau$ from the offline dataset and processing them through the encoder to get a basic encoding $\epsilon$. This encoding passes through a causal layer to produce the causal task representation $z$. 
To train the encoder, $z$ is used in the decoder to reconstruct the task context, calculating the causal representation learning loss. We enhance the encoder's discriminability for unseen tasks by introducing mutual information optimization with negative samples $z^*$ from random noise and contrastive learning using real task representations as negative samples for contrastive loss. 

In the meta-training module: we sample context trajectories $\tau$ from the offline dataset, using the trained causal encoder to derive the causal task representation $z$. This representation, along with $\tau$, helps generate the policy $\pi(a|s, z)$ and the action value $Q(s, a, z)$ via policy and action networks. The policy loss ($L_{actor}$) and action value loss ($L_{critic}$) are computed using SAC and used to update the networks accordingly. 
In the meta-testing module: The learned policy and action value networks are fixed, and their performance is evaluated on test tasks to assess the effectiveness of the causal task encoder and policy. These modules work together to effectively leverage causal representation, enhancing generalization and performance in context-based OMRL.

\subsection{Causal Task Encoder with Causal Representation Learning}
In order to construct a causal task representation, we adopt a causal variational autoencoder~\cite{kingma2013auto}, as it effectively handles the randomness in transition trajectory caused by stochastic policy. 
As shown in Figure \ref{framework}, the causal autoencoder consists of three components: an encoder $\mathrm{E}$, a causal layer $\mathrm{C}$, and a decoder $\mathrm{D}$.
In causal autoencoder: task context \(\tau=\{(s_{j}, a_{j}, s'_{j}, r_{j}, \eta_{j})\}^{t}_{j=t-n+1}\) is the observed variable; $z \in \mathbb{R}^n$ and $\epsilon \in \mathbb{R}^n $ are latent variables. 
To better learn the causal autoencoder, we introduce the additional information $u=(g_{\mathrm{info}}, s_{\mathrm{info}}, a_{\mathrm{info}}, r_{\mathrm{info}})$, where $g_{\mathrm{info}}$ denotes the task goal,  
$s_{\mathrm{info}}$, $a_{\mathrm{info}}$, and $r_{\mathrm{info}}$ denote state, action, and reward information, respectively. $g_{\mathrm{info}}, s_{\mathrm{info}}, a_{\mathrm{info}},$ and $  r_{\mathrm{info}}$ are separately sampled from the Gaussian distribution defined by the mean and variance of their corresponding values in $\tau$. 

\noindent\textbf{\emph{Encoder, Causal Structure, and Latent Distribution.}} 
During the inference, the encoding process in our model can be represented as:
\begin{flalign}
    \epsilon = \mathrm{E}(\tau,u) + \xi,
\end{flalign}
where $\xi$ is a vector of independent noise characterized by a probability density $p_{\xi}$. 

The causal layer $\mathrm{C}$ implements a linear structural causal model~\cite{shimizu2006linear} 
\begin{flalign}\label{task_representation_z}
	z  = ( \mathrm{I} - \mathrm{A}^T)^{-1}\epsilon,
\end{flalign}
where $\mathrm{A}$ represents the causal graph $\mathrm{A}=[\mathrm{A}_1|\dots|\mathrm{A}_n]$, with each $\mathrm{A}_i \in \mathbb{R}^n $ being a weight vector. The element $\mathrm{A}_{ji}$ encodes the causal strength from $z_j$ to $z_i$.

The causal variational autoencoder defines both the posterior distribution and the prior distribution for the latent variables $z$ and $\epsilon$:
\begin{itemize}
    \item \textbf{Posterior Distribution:} The posterior distribution, conditioned on the task context $\tau$ and additional information $u$, is given by:
    \begin{flalign}
    q_{\phi}(z,\epsilon | \tau, u) = q_{\phi}(z|\epsilon) q_{\phi}(\epsilon|\tau, u), \label{causal_encoder_form}
\end{flalign}
where $\phi=(\mathrm{E},\mathrm{A})$ includes the parameters of the encoder and causal layer.
    \item \textbf{Prior Distribution:} The joint prior distribution over the latent variables $z$ and $\epsilon$ is defined by the model parameters $\theta= (\mathrm{E},\mathrm{D},\mathrm{A}, \mathbf{\lambda})$, where $\mathbf{\lambda}$ denotes the parameters of prior distribution for $z$ and $\epsilon$:
    \begin{flalign}
    p_{\theta}(z, \epsilon|u) = p_{\epsilon}(\epsilon)p_{\theta}(z|u),
\end{flalign}
where $p_{\epsilon}(\epsilon)= \mathcal{N}(\mathbf{0}, \mathbf{\mathrm{I}})$, and $p_{\theta}(z|u)$ denotes a factorized Gaussian distribution:
\begin{flalign}
    p_{\theta}(z|u) = \prod_{i=1}^n p_{\theta}(z_i|u_i),
\end{flalign}
with $p_{\theta}(z_i|u_i) = \mathcal{N}(\lambda_1(u_i), \lambda_2^2(u_i) )$, where $\lambda_1$ and $\lambda_2$ are arbitrary functions. 
\end{itemize}

\noindent\textbf{\emph{Decoding Process.}}
The decoding process reconstructs the task context $\tau$ from the latent $z$ as:
\begin{flalign}
    \tau = \mathrm{D}(z) + \zeta,
\end{flalign}
where $\zeta$ is independent noise with a probability density $q_{\zeta}$.
The decoder model is defined as $p_{\theta}(\tau|z,\epsilon, u)$. 
For a task context $\tau$, the conditional generative model parameterized by $\theta$ is:
\begin{flalign}
    p_{\theta}(\tau,z, \epsilon|u) = p_{\theta}(\tau|z,\epsilon,u) p_{\theta}(\epsilon,z|u).
\end{flalign}

\noindent\textbf{\emph{Learning Causal Autoencoder.}}
The learning process involves approximating the true posterior distribution $p_{\theta}(\epsilon, z | \tau, u)$ with a tractable distribution $q_{\phi}(\epsilon, z | \tau, u)$. This is achieved by maximizing the evidence lower bound (ELBO)~\cite{kingma2013auto}:
\begin{flalign}
\begin{aligned}
   & \mathbb{E}_{q_{\tau}}[\mathrm{log}p_{\theta}(\tau|u)] \geq \\
    &\underbrace{ \mathbb{E}_{q_{\tau}}\Big[\mathbb{E}_{\epsilon,z \sim q_{\phi}}[\mathrm{log}p_{\theta}(\tau|z,\epsilon,u)] 
     - \mathrm{KL}(q_{\phi}(\epsilon,z|\tau,u) || p_{\theta}(\epsilon,z|u) )  \Big]}_{\mathrm{ELBO}},\label{elbo}
\end{aligned}
\end{flalign}
where $q_{\tau}$ denotes the data distribution over task context $\tau$ set. We denote the ELBO as $\mathcal{L}_{\mathrm{ELBO}}$. 

To effectively learn the causal graph $\mathrm{A}$, the following constraints are adopted.

\begin{itemize}
    \item \textbf{Task Information Constraint:} To ensure that different tasks correspond to different causal graphs, we applied the following constraints on the causal graph $\mathrm{A}$ using task information $u$~\cite{vowels2022d}:
\begin{flalign}
  \mathcal{L}_{\mathrm{u}} = \mathbb{E}_{\mathcal{D}} \| u - \sigma(\mathrm{A}^Tu) \|_2^2 \leq k_1, \label{c_1}
\end{flalign}
where $\sigma$ denotes the logistic function and $k_1$ is a small positive constant value. 

\item \textbf{DAG Constraint:} This ensures that the causal adjacency matrix $\mathrm{A}$ forms a directed acyclic graph (DAG)~\cite{yu2019dag}:
\begin{flalign}
\mathcal{L}_{\mathrm{A}} \equiv {tr}\left((I + \frac{c}{n} \mathrm{A} \circ \mathrm{A})^n\right) - n =0, \label{c_3}
\end{flalign}
where $c$ denotes an arbitrary value, $\circ$ denotes element-wise multiplication. 
\end{itemize}

The optimization objective for training the causal variational autoencoder is then formulated as a constrained optimization problem:
\begin{flalign}
\begin{aligned}
      &\max \mathcal{L}_{\mathrm{ELBO}} \\
    & \text{s.t.} (\ref{c_1}) (\ref{c_3}).  
\end{aligned}
\end{flalign}

According to the Lagrangian multiplier method, the causal autoencoder can be learned by minimizing $\mathcal{L}_{\mathrm{causal}} $:
\begin{equation}\label{causal_eq}
	\mathcal{L}_{\mathrm{causal}} = - \mathcal{L}_{\mathrm{ELBO}} + \alpha \mathcal{L}_\mathrm{A} + \beta \mathcal{L}_u ,
\end{equation}
where \( \alpha \) and \( \beta \) denote hyperparameters.

\subsection{Enhancing Causal Task Encoder with Mutual Information and Contrastive Learning}

\subsubsection{Mutual Information Optimization}
To better learn task representation, we utilize mutual information to further optimize the causal autoencoder. 
The causal encoder $q_{\phi}$ can be modeled as: $z \sim q_{\phi}(z|\tau)$. 
The task $M$ is sampled from the task distribution $P(M)$, and the distribution of $\tau$ is influenced by both $M$ and the behavior policy. The goal of training the task causal encoder is to maximize the following objective:
\begin{equation}\label{mutual_information_objective}
	\max I(z;M) =\mathbb{E}_{M,z} \left[ \log\frac{p(M|z)}{p(M)} \right].
\end{equation}
It is intractable to optimize the above mutual information in practice. 
Inspired by the InfoNCE~\cite{oord2018representation}, we derive a lower bound for Eq. (\ref{mutual_information_objective}).

\begin{statement}
    Consider a set of tasks $\mathcal{M}$ sampled from a task distribution, where $|\mathcal{M}| =N$. Let $M=\left(S, A, P, R, \rho\right) \in \mathcal{M}$ be the first task, with $\tau$ as its task context, and assume $z\sim q_{\phi}(z|\tau)$. 
    Define the function $f(\tau,z) =\frac{q_{\phi}(z|\tau)}{q_{\phi}(z)}$. 
     For any task $M^*=\left(S^*, A^*, P^*, R^*, \rho^*\right) \in \mathcal{M}$, let $\tau^*$ represent the task context of task $M^*$. Then, the following inequality holds:
    \begin{flalign}
    \label{eq_lower_bound}
	I(z;M) \geq I_{\mathrm{NCE}}(z, M) =\mathbb{E}_{M,z,\tau} \left[ \log \left( \frac{f(\tau, z)}{\sum_{M^* \in \mathcal{M}} f(\tau^*, z)} \right) \right],
\end{flalign}
\end{statement}
The derivation of Eq. (\ref{eq_lower_bound}) is detailed in Appendix A.1.

For simplicity, we use cosine similarity $C(z,z')$ to approximate $f(\tau, z)$ in Eq. (\ref{eq_lower_bound}), where $z$ and $z'$ originate from the same task $M$. Then, the mutual information optimization can be expressed as minimizing $\mathcal{L}_{\mathrm{info}}$:
\begin{flalign}
\mathcal{L}_{\mathrm{info}} = -\mathbb{E}_{M,z,\tau} \left[\log \frac{\exp(C(z,z'))}{\sum_{{M_i}^* \in \mathcal{M}} \exp(C(z,z^*))}\right], \label{info_loss}
\end{flalign}
where $z^*\sim q_{\phi}(z^*|\tau^*)$, $\tau^*$ is generated by adding random noise obeying the Gaussian distribution to $\tau$.

\subsubsection{Contrastive Learning}
To enhance the discriminability of task representations, we use metric-based contrastive learning with Euclidean distance, as defined by the following triplet loss~\cite{li2020multi}:
\begin{flalign}
\label{triplet_loss}
\begin{aligned}
\mathcal{L}_{\mathrm{triplet}} = \sum_{\substack{M_i, M_j \in \mathcal{M} \\ j=i}} \sum_{\substack{M_i,M_k \in \mathcal{M}  \\ k \neq i}} \max(0, ||z_i - z_j||_2^2 - ||z_i - z_k||_2^2 + m),
\end{aligned}
\end{flalign}
where $m = e^{-||z_i - z_j||_2^2}$ represents an adaptive threshold based on the similarity between positive and negative samples.

We also introduce hard sample mining~\cite{shrivastava2016training}, where the closest negative sample becomes the hardest negative sample, and its distance loss is minimized:

\begin{equation}\label{hard_loss}
	\mathcal{L}_{\mathrm{hardest}} = \frac{1}{\max_{M_i,M_k \in \mathcal{M} , i \neq k}(||z_i - z_k||_2^2) + \varsigma},
\end{equation}
where $\varsigma$ is a small constant. Combining the triplet loss and hard sampling mining loss, we obtain the contrastive learning loss:
\begin{equation}\label{contrastive_loss}
	\mathcal{L}_{\mathrm{contrastive}} = \delta \cdot \mathcal{L}_{\mathrm{triplet}} + \kappa \cdot L_{ \mathrm{hardest}}, 
\end{equation}
where \( \delta \) and \( \kappa \) denote hyperparameters.
We combine the mutual information optimization loss and contrastive learning to further optimize the causal task encoder:
\begin{equation}
	\label{loss2}
	\mathcal{L}_{\mathrm{combine}} = \mathcal{L}_{\mathrm{contrastive}} + \nu \cdot \mathcal{L}_{\mathrm{info}}
\end{equation}
where \( \nu \) denotes a hyperparameter.

\subsection{Practical Implementations of CausalCOMRL}
We briefly summarize the practical implementations of CausalCOMRL. 
For the causal autoencoder training process, we begin by sampling task contexts $\tau$ and $\tau^{\prime}$ from datasets $\mathcal{D}$. 
We then compute task representation $z$ and $z^{\prime}$ according to Eq. (\ref{task_representation_z}). 
Next, we construct the loss functions $\mathcal{L}_{\mathrm{causal}}$ and $\mathcal{L}_{\mathrm{combine}}$ to train the causal autoencoder. 
During the meta-training phase, the causal task encoder remains fixed while the policy network \( \pi_{\psi}(a|s, z) \) is trained using data from offline datasets.
For the meta-testing process, we first sample a test task $M$ and its corresponding task context $\tau$. 
The task representation $z$ is then obtained using the causal task encoder. Based on this representation, the agent takes actions according to policy \( \pi_{\psi}(a|s, z) \) and interacts with the environment. 
The detailed pseudo-code of CausalCOMRL can be found in Algorithm~\ref{train} and~\ref{test}. 

\begin{algorithm}[htb!]  
	\caption{Causal Task Encoder Training and Meta-Training}  
	\label{train}
	\begin{algorithmic}[1]
		\REQUIRE
		Datasets $\mathcal{D}$; inference model with parameters $\phi=(\mathrm{E},\mathrm{A})$, generative model with parameters $\theta= (\mathrm{E},\mathrm{D},\mathrm{A}, \mathbf{\lambda})$; RL model $ Q_{\omega}(s,a,z)$, $ \pi_{\psi}(a|s,z)$.
		\STATE\textbf{A. Training causal task encoder: } 
		\REPEAT
		\STATE Sample a train task $M$ and two transition context trajectory $\tau, \tau'$ from $\mathcal{D}$.
		\STATE Obtain task representation $z$ and $z^\prime$ according to Eq. (\ref{task_representation_z}).
		\FOR{$M^* \in \mathcal{M}$}
		\STATE Add noise to $\tau$ to get $\tau^*$.
		\STATE Obtain task representation $z^*$ according to Eq. (\ref{task_representation_z}). 
		\ENDFOR
		\STATE Compute $\mathcal{L}=\mathcal{L}_{\mathrm{causal}} + \mathcal{L}_{\mathrm{combine}}$ in Eq.(\ref{causal_eq}) and Eq.(\ref{loss2}).
		\STATE Update $\phi$ and $\theta$ to minimize $\mathcal{L}$.
		\UNTIL{Done}
		
		\STATE \textbf{B. Train the policy: }
		\REPEAT
		\STATE Sample a train task dataset $M$ and a context trajectory $\tau$ from $\mathcal{D}$.
		\STATE Obtain task representation $z$ according to Eq. (\ref{task_representation_z}). 
		\STATE Update $\omega, \psi$ with offline RL method (SAC). 
		\UNTIL{Done}
	\end{algorithmic}
\end{algorithm}

\begin{algorithm}[htb!]  
	\caption{Meta-Testing}  
	\label{test}
	\begin{algorithmic}[1]
		\REQUIRE
		Datasets $\mathcal{D}$; inference model with parameters $\phi=(\mathrm{E},\mathrm{A})$; RL policy model $ \pi_{\psi}(a|s,z)$.
		\STATE Sample a test task $M$ and a context trajectory $\tau$ from $\mathcal{D}$.
		\STATE Obtain task representation $z$ according to Eq. (\ref{task_representation_z}). 
		\REPEAT
		\STATE Observe $s$, execute $a \sim \pi_{\psi}(a|s, z)$, get $r$ and $s'$.
		\UNTIL{Environment terminates}
	\end{algorithmic}
\end{algorithm}

\section{Experiments}
In this section, we evaluate the proposed CausalCOMRL on presentative meta-RL benchmarks. We first detail our experimental settings, including benchmarks, offline data collection, and context-based OMRL methods for comparison. 
We then evaluate CausalCOMRL against representative context-based OMRL methods and visualize the task representations by projecting the embedding vectors into 2D space for qualitative analysis. 
Finally, we conduct the ablation study to validate the key components of the task encoder in CausalCOMRL.

\subsection{Experimental Settings}
In this section, we describe the experimental settings from three aspects: meta-RL benchmarks, offline data collection strategy, and compared methods. We use four multi-task MuJoCo representative benchmarks~\cite{todorov2012mujoco}: (1) \textbf{Half-Cheetah-Vel}, targeting different velocities for a cheetah-like agent; (2) \textbf{Ant-Dir}, aiming to direct an ant-like agent towards various target directions; (3) \textbf{Hopper-Rand-Params}, focusing on maximizing forward velocity; (4) \textbf{Walker-Rand-Params}, where a biped agent must adapt to forward motion. The first two benchmarks vary in reward functions and transition dynamics, while the last two vary only in transition dynamics.
During the offline data collection, we sample $20$ training tasks and $20$ testing tasks from task distribution for each meta-RL benchmark. For each training task, we adopt SAC~\cite{haarnoja2018soft} to train a a single-task policy. The trajectories in the replay buffer are collected to create the offline datasets. Next, we briefly introduce the several representative context-based OMRL methods used in experiments: Offline-PEARL~\cite{rakelly2019efficient}, FOCAL~\cite{li2020focal}, and CORRO~\cite{yuan2022robust}. Specifically, Offline PEARL extends PEARL to offline reinforcement learning. FOCAL and CORRO separately utilize metric learning and contrastive learning to learn task encoder.

\subsection{Comparison with State-of-the-art Methods}
\begin{figure*}[htb!]
	\centering
	\includegraphics[width=\textwidth]{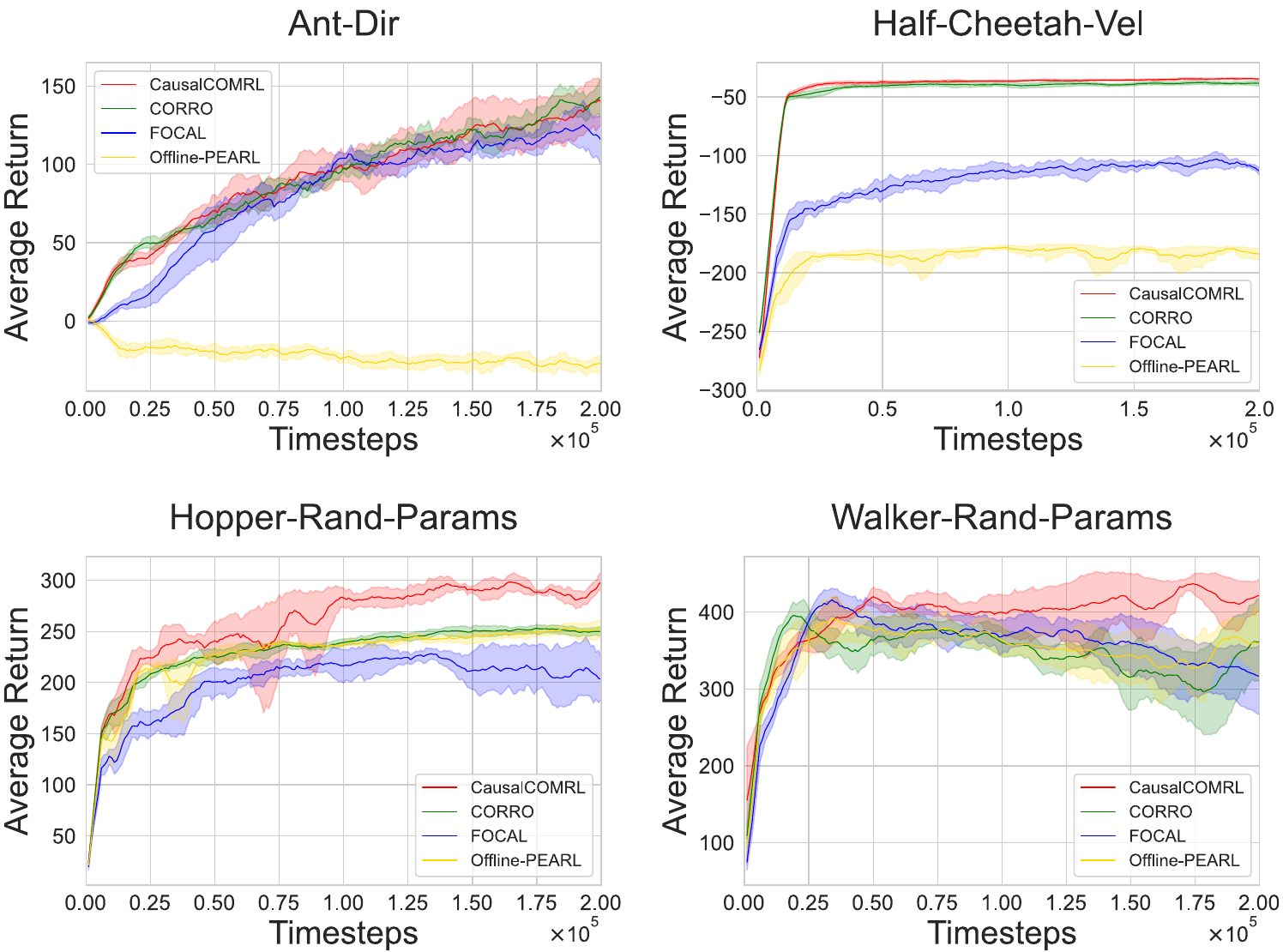}
	\caption{Average test returns of CausalCOMRL against representative context-based OMRL methods on four environments in out-of-distribution tasks. The $X$-axis and $Y$-axis denote the timesteps and average return, respectively. The shaded region shows standard deviation across 5 seeds.}
	\label{result_method}
\end{figure*}

To evaluate the proposed CausalCOMRL, we compare CausalCOMRL against other representative context-based COMRL methods on four meta-RL benchmarks. 
Figure \ref{result_method} shows the performance comparison in out-of-distribution tasks. As shown in Figure \ref{result_method}, we observe that CausalCOMRL achieves higher or comparable final returns than other methods (CORRO, FOCAL, and Offline-PEARL) in most environments. The gap between the returns of CausalCOMRL on Hopper-Rand-Params and those of other methods is particularly notable. 

Table \ref{tab:final} presents the results of the compared methods, evaluated at the final training epoch. In this table, `IID' refers to the in-distribution training tasks, while `OOD' refers to the out-of-distribution tasks. As shown, CausalCOMRL achieves the highest returns among the methods (CORRO, FOCAL, Offline-PEARL, and CausalCOMRL) on both Hopper-Rand-Params and Walker-Rand-Params.
For the OOD tasks, our method outperforms others on most tasks, including Half-Cheetah-Vel, Hopper-Rand-Params, and Walker-Rand-Params. The only exception is Ant-Dir, where the return is slightly lower than CORRO.
For IID tasks, CausalCOMRL delivers the highest returns on three tasks (Ant-Dir, Hopper-Rand-Params, and Walker-Rand-Params) with only a small gap observed on Half-Cheetah-Vel. 
The results from Figure \ref{result_method} and Table \ref{tab:final} demonstrate that the proposed CausalCOMRL outperforms other context-based OMRL methods on most environments. 

\begin{table}[htbp]
\centering
\resizebox{\textwidth}{!}{
\begin{tabular}{lccccccccc}
\toprule
Method & & Ant-Dir & Half-Cheetah-Vel & Hopper-Rand-Params & Walker-Rand-Params \\
\midrule
\multirow{2}{*}{Offline PEARL} & IID & $14.9\pm5.2$ & $-168.1\pm10.4$ & $265.0\pm17.3$ & $356.1\pm64.6$ \\
& OOD & $-23.7\pm10.9$ & $-183.8\pm6.3$ & $255.9\pm13.2$ & $358.4\pm61.0$ \\
\midrule
\multirow{2}{*}{FOCAL} & IID & $169.7\pm22.2$ & $-112.0\pm9.8$ & $213.7\pm23.0$ & $312.4\pm38.5$ \\
& OOD & $112.0\pm29.8$ & $-115.3\pm8.2$ & $196.5\pm21.6$ & $318.6\pm57.2$ \\
\midrule
\multirow{2}{*}{CORRO} & IID & $235.7\pm26.4$ & $\mathbf{-33.2\pm1.5}$ & $277.1\pm5.5$ & $355.1\pm78.2$ \\
& OOD & $\mathbf{146.9\pm23.9}$ & $-37.7\pm3.0$ & $250.8\pm7.4$ & $368.1\pm66.4$ \\
\midrule
\multirow{2}{*}{CausalCOMRL} & IID & $\mathbf{246.8\pm24.6}$ & $-37.9\pm2.4$ & $\mathbf{291.7\pm15.7}$ & $\mathbf{448.5\pm17.3}$ \\
& OOD & $141.6\pm17.7$ & $\mathbf{-35.0\pm1.2}$ & $\mathbf{305.8\pm13.4}$ & $\mathbf{426.7\pm26.3}$ \\
\bottomrule
\end{tabular}
}
\caption{Average test returns of different methods in IID and OOD settings for various environments. IID means the in-distribution training tasks, while OOD means the out-of-distribution tasks. The best results for each environment are highlighted in bold.}
\label{tab:final}
\end{table}

\subsection{Latent Space Visualization}
\begin{figure*}[htb!]
	\centering
	\includegraphics[width=0.95\textwidth]{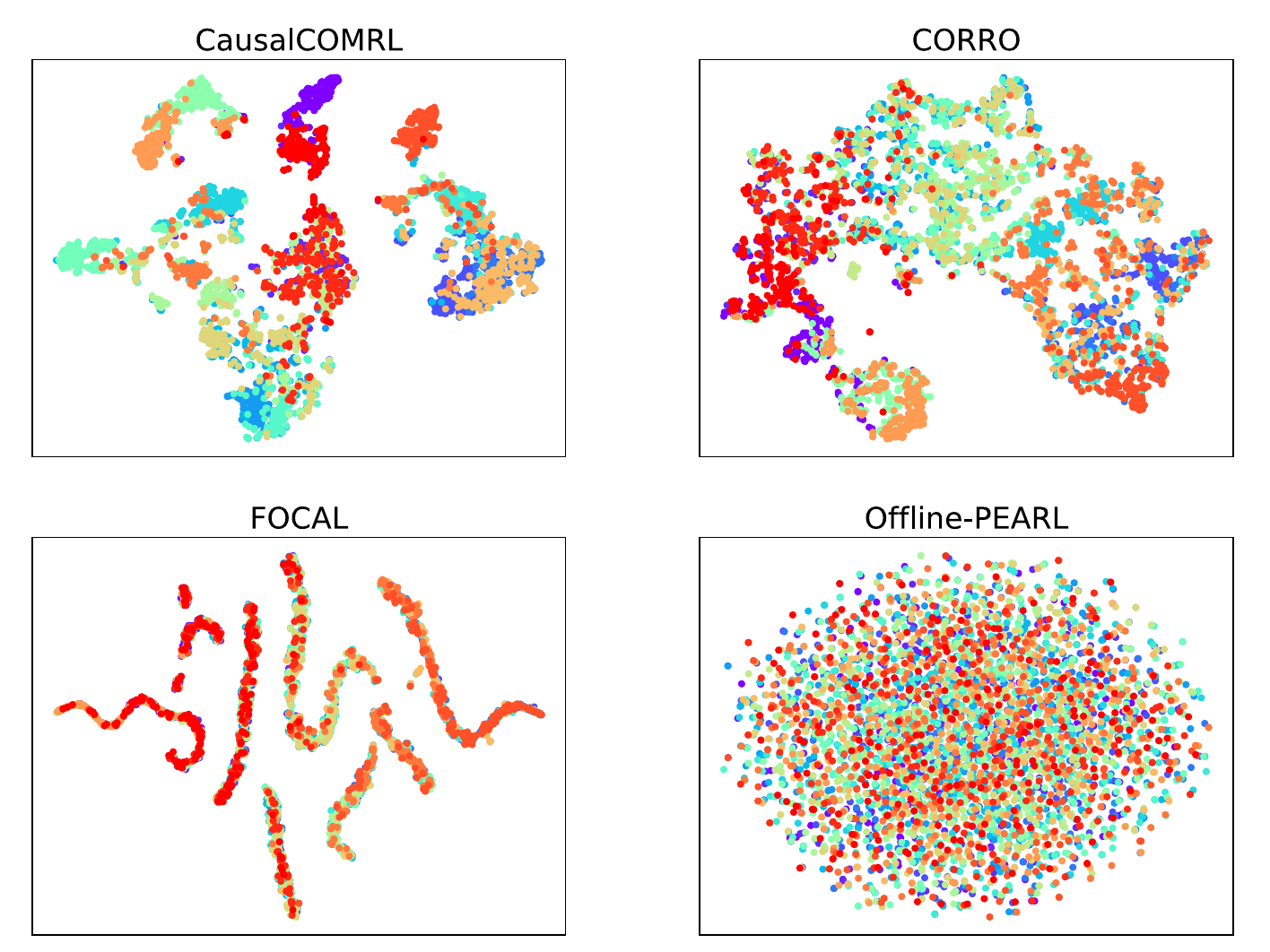}
	\caption{The t-SNE visualization of the task representation space in Half-Cheetah-Vel. Test task points are uniformly sampled from test tasks and color-coded from red to purple for velocities 0 to 3.}
	\label{visualization}
\end{figure*}

To analyze the task representation produced by the context encoder, we utilize t-SNE\footnote{For fair comparison, we use the t-SNE code provided by the official sklearn library (\url{https://scikit-learn.org/}).}~\cite{van2008visualizing} to translate embedded vectors into a 2D space for visualizing task representations. We visualize the task embeddings by sampling 200 transitions from the relay buffer for every test task in HalfCheetah-Vel. The comparison for task representation visualization is presented in Figure \ref{visualization}. 
The visualization result demonstrates that CausalCOMRL provides more distinct contrasts and clearer boundaries between transitions compared to other methods (PEARL, FOCAL,and CORRO). This indicates that CausalCOMRL more effectively distinguishes transitions associated with different tasks, a capability that is attributed to the causal task encoder in CausalCOMRL. The latent space visualization results presented in Figure \ref{visualization} validate the effectiveness of the causal task encoder in CausalCOMRL.

\subsection{Ablation Study}
The learning of task context encoder is crucial in CausalCOMRL, enhancing task representation and distinctiveness, which supports meta-policy learning. We improve the encoder by integrating causal representation learning and introducing a hybrid loss combining mutual information optimization and contrastive learning (see Eq. (\ref{loss2})). To assess these enhancements, we conducted ablation experiments, with results shown in Table~\ref{ablation}.
The `Encoder' in Table~\ref{ablation} is the encoder component of the standard autoencoder.

\begin{table*}[!htb!]
	\centering
	\resizebox{\textwidth}{!}{
		\begin{tabular}{ccccc}
			\toprule
			Method  & Hopper-Rand-Param & Walker-Rand-Param & Ant-Dir & Half-Cheetah-Vel \\
			\hline \hline
			Causal Encoder + $\mathcal{L}_{\mathrm{combine}}$ & $305.8\pm13.4$ & $\mathbf{426.7\pm26.3}$ & $\mathbf{141.6\pm17.7}$ & $\mathbf{-35.0\pm1.2}$ \\
			Encoder + $\mathcal{L}_{\mathrm{combine}}$ & $\mathbf{310.7\pm15.3}$ & $365.0\pm86.0$ & $112.6\pm24.3$ & $-52.2\pm3.7$ \\
			Causal Encoder & $297.3\pm23.5$ & $416.0\pm14.7$ & $90.0\pm13.4$ & $-36.2\pm2.0$ \\
			Encoder & $285.6\pm20.4$ & $421.6\pm30.1$ & $-3.1\pm14.1$ & $-89.2\pm8.2$ \\
			\bottomrule
		\end{tabular}
	}
 	\caption{Ablation study on task context encoder methods in out-of-distribution tasks. The best results are denoted in bold. The `Encoder' is the encoder component of the standard autoencoder.}
	\label{ablation}
\end{table*}

In Table~\ref{ablation}, it is evident that the causal encoder method with the combined loss delivers higher return values than the causal encoder without the combined loss approach. Similarly, the encoder method with the combined loss outperforms the standard encoder method, indicating that mutual information optimization and contrastive learning substantially improve task encoder learning. Furthermore, in most environments, the returns for the causal encoder method surpass those of the encoder method and this trend continues when comparing causal encoder with the combined loss to encoder with the combined loss. This highlights the impact of causal representation learning in enhancing the task encoders.

\section{Conclusion}
In this paper, we introduce CausalCOMRL, a context-based OMRL method that employs causal representation learning to enhance task representation and generalization. We utilize a causal variational autoencoder to incorporate causal relationships into task representations. To better learn task representation, we integrate mutual information optimization and contrastive learning into the task encoder. 
Using this enhanced task representation, we apply SAC algorithm for meta-policy learning during meta-training. 
The extensive experiments demonstrate that CausalCOMRL outperforms other context-based OMRL methods in most cases, confirming the effectiveness of our causal model. 

\section*{CRediT authorship contribution statement}
\textbf{Zhengzhe Zhang}: Writing – review \& editing, Writing – original draft, Visualization, Validation, Software, Methodology, Conceptualization.
\textbf{Wenjia Meng}: Writing – review \& editing, Writing – original draft, Supervision, Resources, Investigation.
\textbf{Haoliang Sun}: Writing - Review \& Editing, Investigation, Supervision.
\textbf{Gang Pan}: Project administration.

\section*{Declaration of competing interest}
The authors declare that they have no known competing financial interests or personal relationships that could have appeared to influence the work reported in this paper.

\section*{Data availability}
Open data have been used.

\appendix
\section{Derivation of Eq. (\ref{eq_lower_bound})}\label{sec_lower_bound_mutual_information}
According to $I(z;M)$ in Eq. (\ref{mutual_information_objective}), we have
\[
\begin{aligned}
 I(z;M) - \log(N)
& =\mathbb{E}_{M,z} \left[ \log\frac{p(M|z)}{p(M)} \right] - \log(N) \\
& =\mathbb{E}_{M,z} \left[ \log\frac{p(z|M)}{p(z)} \right] - \log(N) \\
& =\mathbb{E}_{M,z} \left[ \log \int_\tau \frac{p(\tau|M)q_\phi(z|\tau)}{q_\phi(z)} d\tau \right] - \log(N) \\
& =\mathbb{E}_{M,z} \left[ \log \mathbb{E}_{\tau} \big[  \frac{q_\phi\left(z | \tau\right)}{q_\phi\left(z\right)}\big]\right] - \log(N) \\
& \geq -\mathbb{E}_{M,z,\tau}\left[ \log \left(\frac{q_\phi\left(z\right)}{q_\phi\left(z | \tau\right)} N\right) \right] \\
& = -\mathbb{E}_{M,z,\tau}\left[ \log \left(1+\frac{q_\phi\left(z\right)}{q_\phi\left(z | \tau \right)}(N-1) \mathbb{E}_{M^* \in \mathcal{M}\backslash \{ M\}} \big[ \frac{q_\phi(z | \tau^*)}{q_\phi(z)} \big]\right) \right]  \\
& =-\mathbb{E}_{M,z,\tau} \log \left[1+\frac{q_\phi\left(z\right)}{q_\phi\left(z | \tau \right)} \sum_{M^* \in \mathcal{M}\backslash \{M\}} \frac{q_\phi\left(z | \tau^*\right)}{q_\phi\left(z\right)}\right] \\
& =\mathbb{E}_{M,z,\tau}\left[\log \frac{\frac{q_\phi\left(z | \tau\right)}{q_\phi\left(z\right)}}{\frac{q_\phi\left(z | \tau\right)}{q_\phi\left(z\right)}+\sum_{M^* \in \mathcal{M}\backslash \{M\}} \frac{q_\phi\left(z | \tau^*\right)}{q_\phi\left(z\right)}}\right] \\
& = \mathbb{E}_{M,z,\tau} \left[ \log \left( \frac{f(\tau, z)}{\sum_{M^* \in \mathcal{M}} f(\tau^*, z)} \right) \right]=I_{\rm{NCE}}(z, M),
\end{aligned}
\]

We can derive $I(z;M) \geq I_{\rm{NCE}}(z, M)$ due to $\log(N)$ is non-negative.

\section{More Experimental Setting}
Our experiments were conducted on a GPU server equipped with four NVIDIA GeForce RTX 3090 GPUs, each with 24GB of memory. For each task, we sampled 40 environments from the environment distribution, allocating 20 environments for training and the remaining 20 for testing. 
Among the hyperparameters used in experiments, those related to the loss function are particularly crucial. 
For these cirtical hyperparameters, we select the optimal values among their respective ranges. 
Their ranges are described as follows: $\alpha \in [1e-1, 1]$, $\beta \in [1e-1, 3]$, $\delta \in [1e-1, 3]$, $\kappa \in [0, 3]$, $\nu \in [1e-1, 3]$.
More hyperparameters and the network settings are detailed in Table~\ref{data collection} and Table~\ref{train_h}.

\begin{table}[htb!]
  \centering
  \resizebox{\textwidth}{!}
  {\begin{tabular}{ccccc}
    \toprule
    Hyperparameters & Ant-Dir & Half-Cheetah-Vel & Hopper-Rand-Params & Walker-Rand-Params \\
    \midrule
    Dataset size & 2e4 & 5e4 & 7e4 & 1e5 \\
    Training steps & 1e6 & 1e6 & 1e6 & 1e6 \\
    Learning rate & 3e-4 & 3e-4 & 3e-4 & 3e-4 \\
    RL network width & 128 & 128 & 128 & 128 \\
    RL network depth & 3 & 3 & 3 & 3 \\
    Batch size & 256 & 256 & 256 & 256 \\
    \bottomrule
  \end{tabular}}
  \caption{Hyperparameters for the collection of the offline dataset.}
  \label{data collection}
\end{table}

\begin{table}[htb!]
  \centering
  \resizebox{\textwidth}{!}
  {\begin{tabular}{ccccc}
    \toprule
    Hyperparameters & Ant-Dir & Half-Cheetah-Vel & Hopper-Rand-Params & Walker-Rand-Params \\
    \midrule
    MLP hidden size & 64 & 64 & 128 & 128 \\
    Latent space dim & 16 & 16 & 16 & 16 \\
    Encoder training steps & 1e6 & 1e6 & 1e6 & 1e6 \\
    Encoder batch size & 256 & 512 & 512 & 256 \\
    Encoder learning rate & 3e-4 & 3e-4 & 3e-4 & 3e-4 \\
    DAG constraint weight $c$ & 4 & 4 & 4 & 4 \\
    Encoder loss weight $\alpha$ & 3e-1 & 3e-1 & 3e-1 & 3e-1 \\
    Encoder loss weight $\beta$ & 1 & 1 & 1 & 1 \\
    Encoder loss weight $\delta$ & 2 & 2 & 2 & 2 \\
    Encoder loss weight $\kappa$ & 2 & 2 & 2 & 2 \\
    Encoder loss weight $\nu$ & 1 & 1 & 1 & 1 \\
    RL training steps & 2e5 & 2e5 & 2e5 & 2e5 \\
    RL network width & 256 & 256 & 256 & 256 \\
    RL network depth & 3 & 3 & 3 & 3 \\
    RL batch size & 256 & 256 & 256 & 256 \\
    RL learning rate & 3e-4 & 3e-4 & 3e-4 & 3e-4 \\
    \bottomrule
  \end{tabular}}
  \caption{Hyperparameters in meta training phases.}
  \label{train_h}
\end{table}

\section{Additional Experimental Results}
\begin{figure}[tpb]
  \centering
  \includegraphics[width=0.85\textwidth]{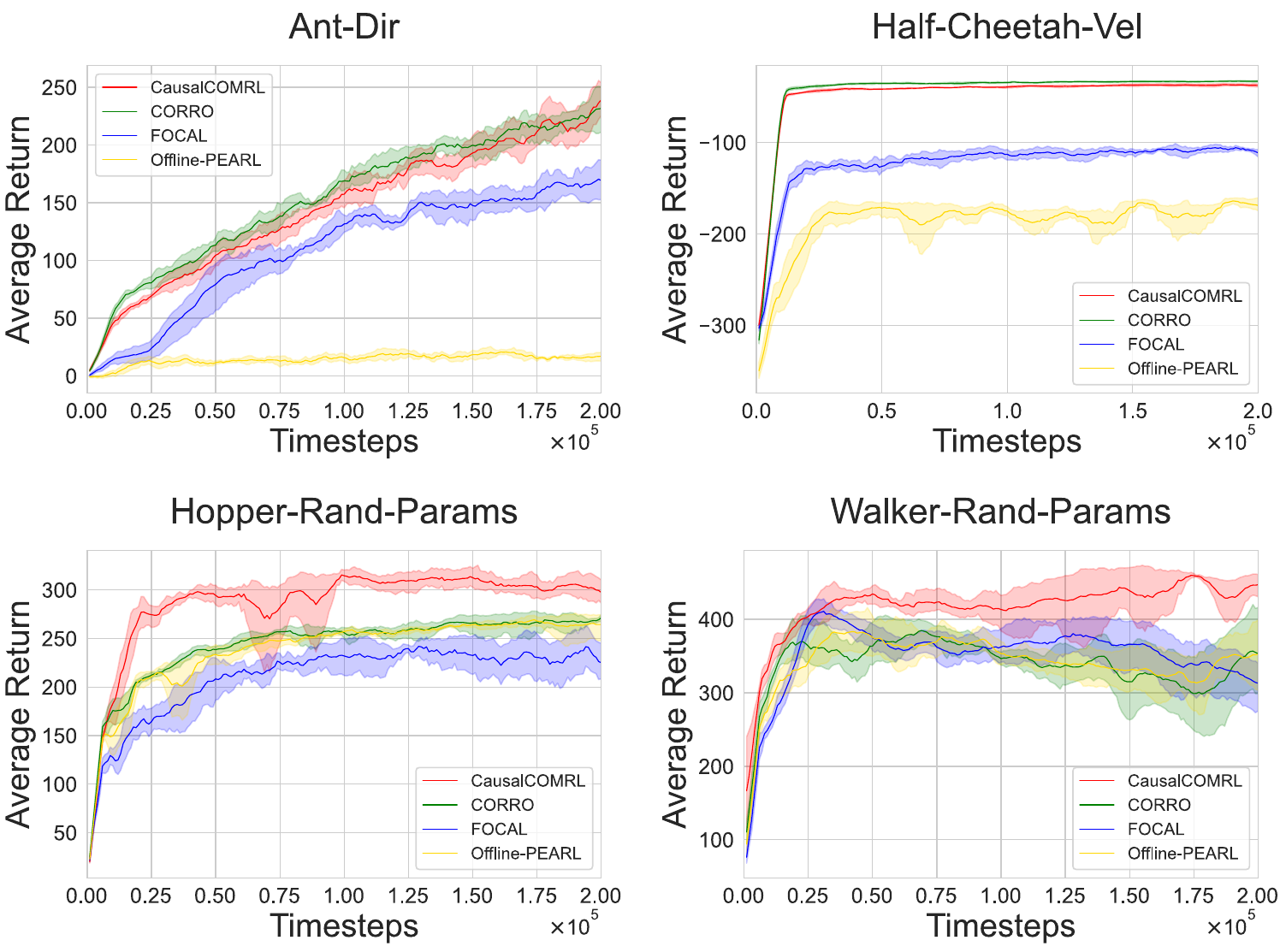}
  \caption{Average returns of CausalCOMRL against the baselines in the in-distribution training tasks. The shaded region shows standard deviation across 5 seeds.}
  \label{iod}
\end{figure}

\noindent\emph{\textbf{Results on Training Tasks.}}
The results of CausalCOMRL compared to other methods across the training tasks of all four environments, using five random seeds, are summarized in Figure \ref{iod}. Performance is evaluated based on the average return across all tasks.

\begin{figure*}[tpb]
  \centering
    \subfloat[Ant-Dir]{
    \parbox{0.5\textwidth}{
    \centering
    \includegraphics[width=0.25\textwidth]{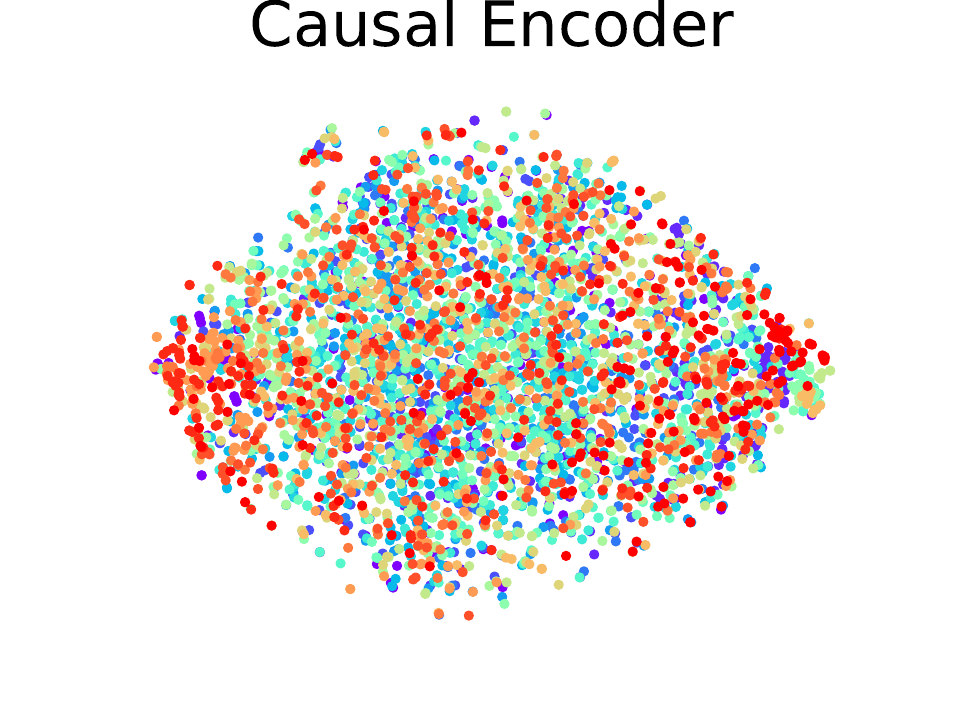}\includegraphics[width=0.25\textwidth]{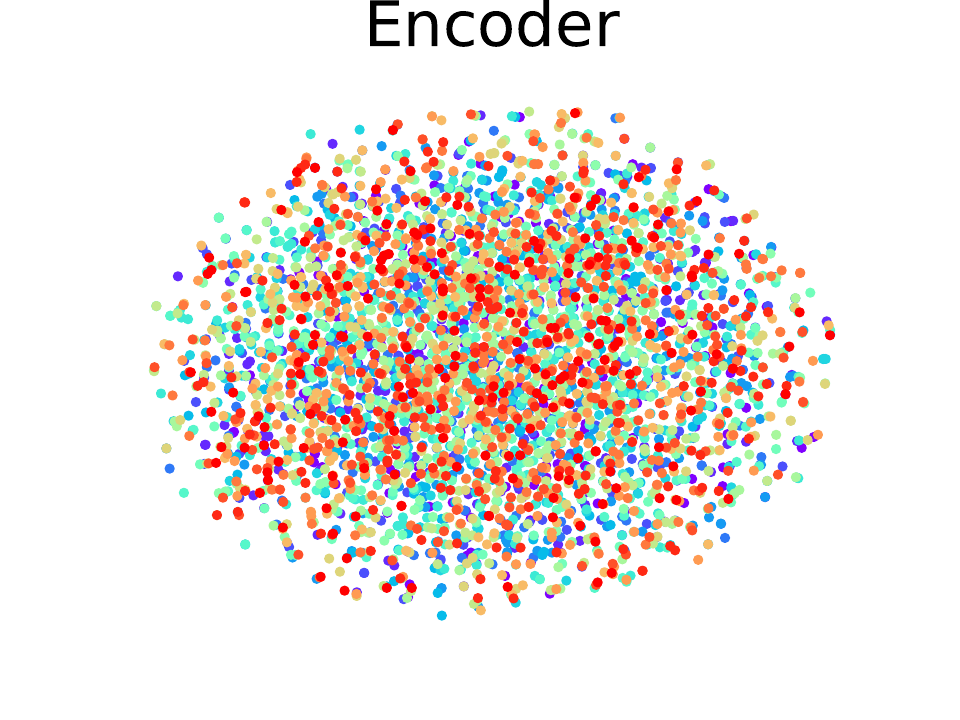}
    }
  }
  \subfloat[Hopper-Rand-Params]{
    \parbox{0.5\textwidth}{
    \centering
    \includegraphics[width=0.25\textwidth]{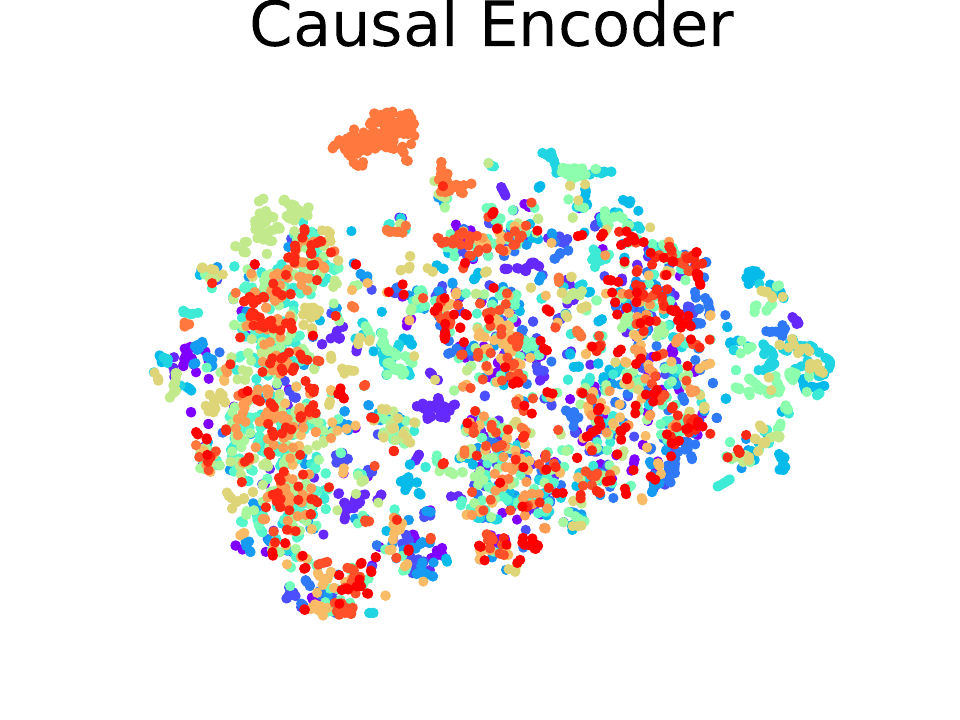}\includegraphics[width=0.25\textwidth]{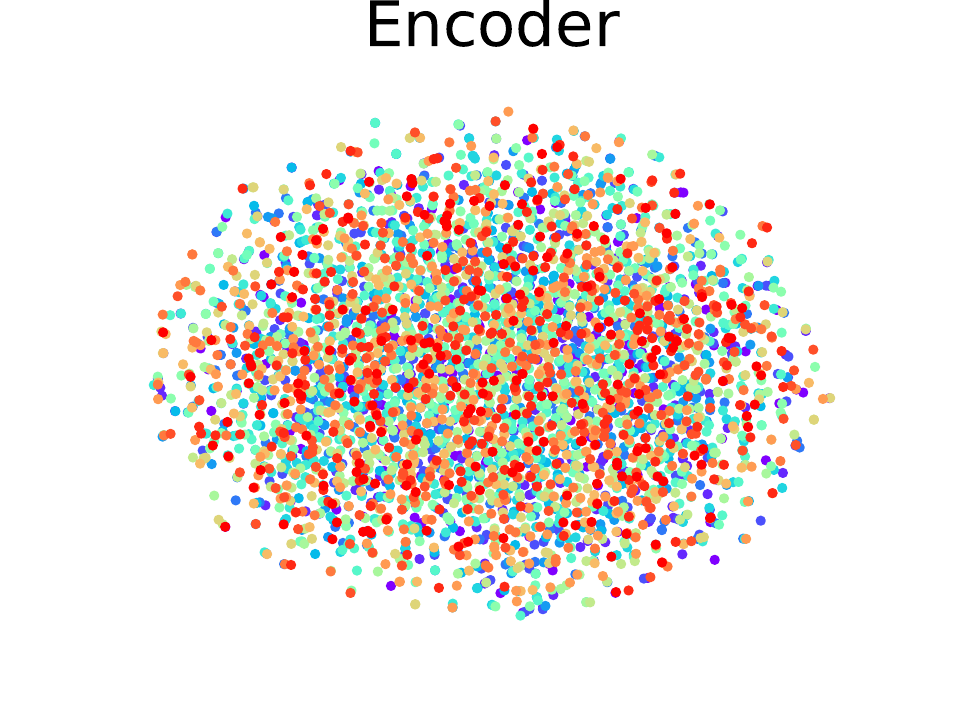}
    }
  }
  \hfill
  \subfloat[Half-Cheetah-Vel]{
    \parbox{\textwidth}{
    \centering
    \includegraphics[width=0.25\textwidth]{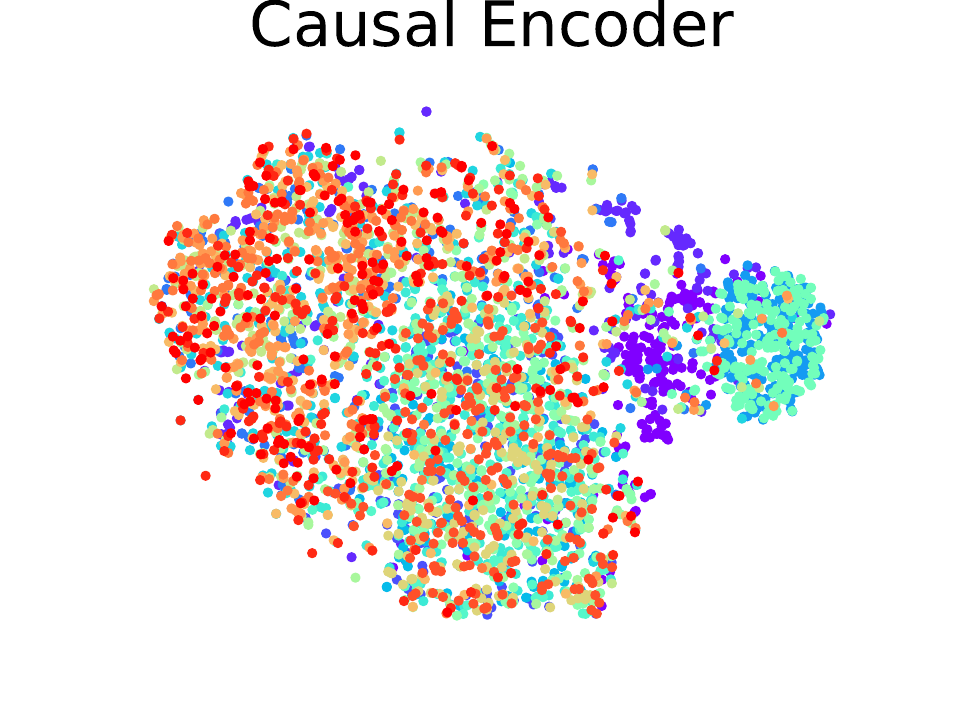}\includegraphics[width=0.25\textwidth]{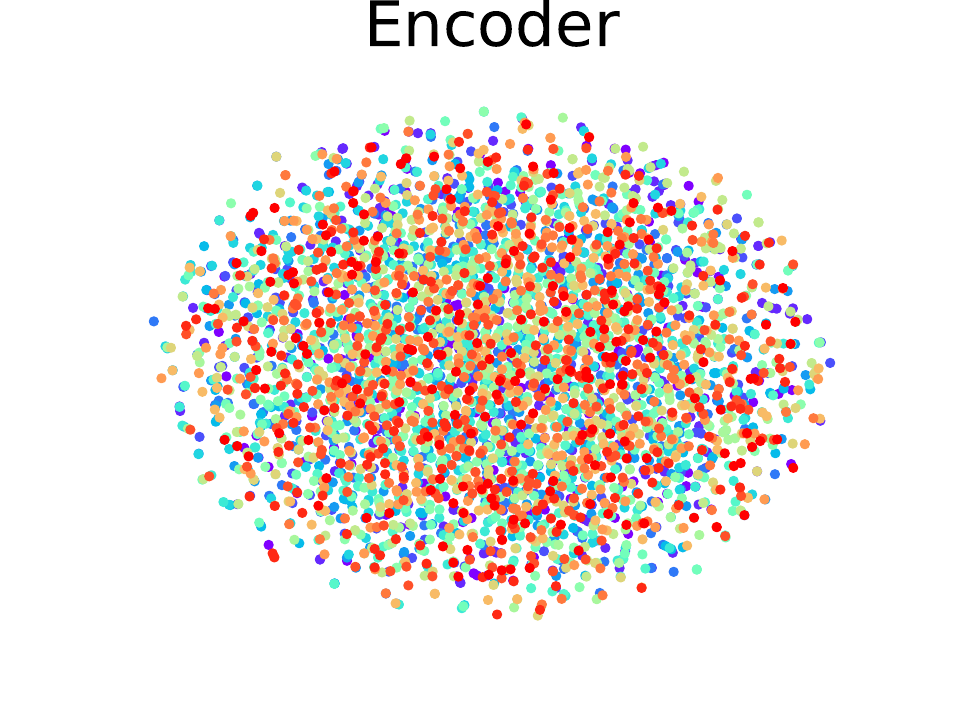}
    }
  }
  \caption{The t-SNE visualizations of the task representation embedding vectors drawn from test tasks on Ant-Dir, Hopper-Rand-Params, and Half-Cheetah-Vel.}
  \label{visual_other}
\end{figure*}

\noindent\emph{\textbf{Additional Results for Task Representation Embedding.}}
Figure \ref{visual_other} provides the t-SNE visualizations of task representation embedding between encoder and causal encoder in Ant-Dir, Hopper-Rand-Params, and Half-Cheetah-Vel environments.

{
    \small
    \bibliographystyle{unsrt}
    \bibliography{elsarticle-num}
}

\end{document}